\documentclass[letterpaper, 10 pt, conference]{ieeeconf}  

\IEEEoverridecommandlockouts                              

\usepackage{graphics} 
\usepackage{epsfig} 
\usepackage{mathptmx} 
\usepackage{times} 
\usepackage{amsmath} 
\usepackage{amssymb}  
\usepackage[dvipsnames]{xcolor}
\usepackage[utf8]{inputenc}
\usepackage{subcaption}
\usepackage{pdfpages}
\usepackage{paralist}
\usepackage[font=small,labelfont=bf]{caption}
\usepackage{array}
\usepackage{booktabs}
\usepackage{lipsum}
\usepackage{caption}
\usepackage{siunitx}
\usepackage{cite}
\usepackage{mathtools}
\usepackage{xcolor}
\usepackage{blkarray, bigstrut}
\usepackage{svg}
\let\labelindent\relax
\usepackage{enumitem}
\usepackage[bookmarksopen, bookmarksdepth=2, breaklinks=true ]{hyperref}
\RequirePackage{color,graphicx}
\definecolor{linkcolour}{rgb}{0.2,0.285,0.918}
\hypersetup{colorlinks,linkcolor={blue},citecolor={linkcolour},urlcolor={blue}} 
\usepackage{graphicx}

\title{{\LARGE \bf \textit{Can a Robot Trust You?\\ }} {\Large A DRL-Based Approach to Trust-Driven Human-Guided Navigation}}
\author{Vishnu Sashank Dorbala, Arjun Srinivasan, and Aniket Bera\\
{University of Maryland, College Park, USA}\\
{\small{Supplemental version including Code, Video, Datasets at \url{https://gamma.umd.edu/robotrust/}}}\vspace{-15pt}

}
\newcolumntype{M}[1]{>{\centering\arraybackslash}m{#1}}

\newcommand\Item[1][]{%
  \ifx\relax#1\relax  \item \else \item[#1] \fi
  \abovedisplayskip=0pt\abovedisplayshortskip=0pt~\vspace*{-\baselineskip}}

\begin{document}

\maketitle
\thispagestyle{empty}
\pagestyle{empty}

\begin{abstract}
Humans are known to construct cognitive maps of their everyday surroundings using a variety of perceptual inputs. 
As such, when a human is asked for directions to a particular location, their \textit{wayfinding} capability in converting this cognitive map into directional instructions is challenged. Owing to spatial anxiety, the language used in the spoken instructions can be vague and often unclear.
To account for this unreliability in navigational guidance, we propose a novel Deep Reinforcement Learning (DRL) based trust-driven robot navigation algorithm that learns humans' trustworthiness to perform a language guided navigation task.

Our approach seeks to answer the question as to whether a robot can trust a human's navigational guidance or not. To this end, we look at training a policy that learns to navigate towards a goal location using only \textit{trustworthy} human guidance, driven by its own \textit{robot trust metric}.
We look at quantifying various \textit{affective} features from language-based instructions and incorporate them into our policy's observation space in the form of a \textit{human trust metric}. We utilize both these trust metrics into an \textit{optimal cognitive reasoning scheme} that decides when and when not to trust the given guidance. Our results show that the learned policy can navigate the environment in an optimal, time-efficient manner as opposed to an explorative approach that performs the same task. We showcase the efficacy of our results both in simulation and a real world environment.



\end{abstract}

\section{Introduction}
 


Recent advances in Robotics and Artificial Intelligence (AI) technologies are gradually enabling humans and robots to team up, co-work, and share spaces in different environments. As humans interact with each other to gain more knowledge about the world, they subconsciously form their individual opinions about each other~\cite{opinion1} along with notions of credibility and trust~\cite{trust1, trust2, trust3}. These notions influence their decision-making capabilities in collaborative environments. In the same way, a social robot's decision-making capabilities also need to account for trust, reliability, and various human behaviors.

Several previous contributions in Human-Robot Interaction (HRI) examine the trustworthiness among humans and AI agents~\cite{hritrust1, hritrust2, hritrust3, hritrust4, hritrust5, hritrust6}. These works mostly address the issue of humans trusting robots and design ``trust metrics" that evaluate if a robot's behavior can be deemed trustworthy. In contrast, the concept of robots questioning human rationality has been less extensively researched. While it is important for the human to trust a robot enough to feel safe about its actions, we feel that it is also important in a collaborative environment for a robot to question human actions so as to execute tasks effectively. 
In this paper, we investigate this notion in a human-guided navigation scenario, where a robot must determine if it can trust the given guidance or not.

\begin{figure}[t]
    \centering
    \includegraphics[width=1\linewidth]{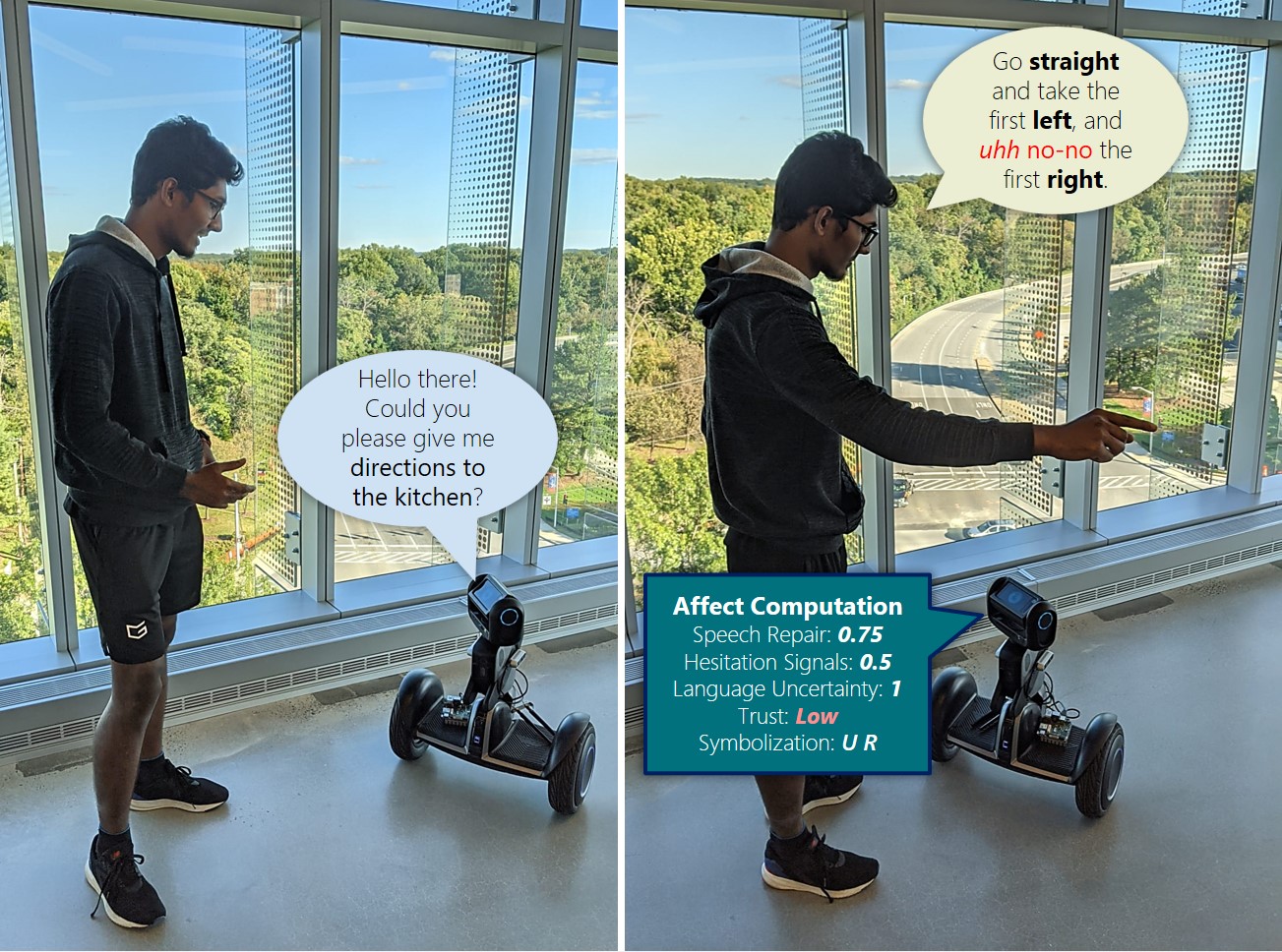}
    \caption{\textit{We look at whether humans can be trusted on the navigational guidance they give to a robot. In order to do this, we first quantify a \textit{human trust metric} using various forms of speech-driven affective cues. We then propose a Deep Reinforcement Learning policy that learns a \textit{robot trust metric} that decides what guidance a robot can trust and what it cannot, based on the environment. Using these two metrics, our algorithm learns to navigate an unseen environment in an optimal time-efficient manner.} \vspace{-0.8cm}}
    \label{fig:Inroduction}
\end{figure}

The use of human guidance for robot navigation has been well-documented in~\cite{dirask1, dirask2, dirask3, dirask4, hrinav}. These methods use information retrieval models along with gesture understanding to guide a robot in an unstructured environment using language commands. Solely using language instructions for robot navigation has also been explored~\cite{langnav0,langnav1, langnav3}. These contributions' primary goal has been to use natural language instructions to guide a robot towards a particular target location in an environment. However, they all work under the assumption that the guiding human knows the target's exact location and perfectly describes instructions to reach that goal, which is not always the case. This is especially true in large, complex environments like malls and theme parks, where people might be new are not well versed with the place.

When humans attempt to give directional commands to a location, their wayfinding and spatial reasoning capabilities come into question \cite{dirsense1, dirsense2, dirsense3, dirsense4}. Furthermore, the spatial anxiety~\cite{spatialnav1, spatialnav2} induced while being asked for directions can lead to errors in judgment. A robot cannot blindly trust human opinion~\cite{trust1,trust3} in this case, and must reason about the guidance it has been given. This aspect of robots \textit{trusting} humans has been explored far less in the literature, motivating us to incorporate the element of robot-human trustworthiness into our work.


To this end, we seek to model this trust and reliability in human guidance from an affective computing standpoint~\cite{cogmap1,reasoning1} and look at how a robot should reason about direction commands that it gets from a human while navigating. 
We look at the design of HRI experiments where the robot's navigational decisions in response to speech-based directional instructions are driven by its \textit{\textbf{gullible} (trusting)} or \textit{\textbf{skeptical} (non-trusting)} personality \cite{persona2, persona4, persona5}. 

We hypothesize that the \textit{optimal} personality that the robot must assume to perform language-guided navigation is very much dependent on the kind of environment in which it is placed. For instance, when placed in an office-like environment where most people know the layout and can be trusted to give precise guidance, a \textit{\textbf{gullible}} personality would interact less, and thus might reach the goal faster. However, in a crowded, mall, or street-like environment, a \textit{\textbf{skeptical}} personality might be more advantageous, as more people might not know the environment well enough to give proper guidance.

In our work, we look at learning this \textit{optimal personality} that a robot must display when placed in an arbitrary environment. To achieve this, we train a \textit{Deep Reinforcement Learning} policy that uses language guidance from bystanders along with an \textit{affective} measure of human trust to learn an adaptive ``\textit{robot trust metric}". Our policy learns to identify and follow only \textit{trustworthy} human guidance to optimize for time-efficient navigation. The key contributions of our work can be summarized as follows:
\begin{enumerate}[nolistsep]
    \item We present a novel Deep Reinforcement Learning (DRL) based approach that gives decisions on whether or not a robot should follow human guidance via a \textit{robot trust metric}. To the best of our knowledge, ours is the first approach to quantify a robot trust metric.
    \item We present an approach to extract \textit{affective} features from human language commands using both text and speech. We later make use of this to model humans in our simulations.
    \item We showcase the practical results of our experimentation on a robot platform in the real world and also present ablation results in a simulation environment.
    \item Finally, we release \textit{Lang2Symb}, a symbolization dataset consisting of 200+ guidance instructions and their corresponding language symbols.
\end{enumerate}
\section{Related Work}
\label{sec:Related}

\subsection{Affective Analysis in Social Robotics}

Social robots need to exist along with humans and interact with them; therefore, understanding the human emotional state plays an important role in their decision making. Emotion Recognition from visual features such a facial expression, gestures , and gaits have been studied in ~\cite{face1,face2,face3,socnav2,gait1,gait2, randhavane2019identifying}. Multimodal and Context-aware models for this task have also been developed recently ~\cite{m3er,multimodal1,emoticon, bhattacharya2020step, mittal2020emotions, bhattacharya2020generating, banerjee2020learning, bera2020you, randhavane2019liar, bhattacharya2019take, randhavane2019eva, randhavane2019modeling}. However, using vision for affective computing in social robots may not be feasible. This is because of the difficulty in obtaining visual features such as facial emotions and gaits that depend on the sensor placement~\cite{affchal1}. Speech is a natural way to communicate with robots and is a practical means of bi-directional communication. Emotion understanding from speech for human-robot interaction has been studied in~\cite{speechemo1,speechemo2}. Bohus et al. and Kennedy et al. \cite{hesitationhri1,hesitationhri2} integrate hesitations and other speech disfluencies into their human-robot interaction system. Eppy et al.~\cite{hesitationhri3} use a cognition model to understand ambiguities in natural language. In our interactive navigation pipeline, direct affective features obtained from disfluencies in speech play an important role.

\subsection{Human-Guided Robot Navigation}

Human language and gestures are natural inputs for communicating with robots and have been studied extensively in HRI. Weiss et al.~\cite{dirask1} and Bauer et al.~\cite{dirask3} conduct autonomous navigation experiments using interactive inputs via language, gestures, and digital interfaces and show that the use of natural language input is an effective and socially acceptable method for navigation. Human gestures and language have also been used together to resolve ambiguities and understand the correct intent of humans in~\cite{srilanka1,fetch-pomdp}. Hu et al.~\cite{Manochanav} describe an approach to follow human instructions in dynamically changing environments making use of semantic maps and language grounding techniques. Gopalan et al. ~\cite{symbols} and Sriram et al.~\cite{iiit} present a method to use language symbols abstractions for robot navigation. We derive our symbolic direction representation using this method. We further build on this work by deducing verbal \textit{affective} features from the language provided and using these features to rectify errors in the given input. End-to-end approaches~\cite{anderson,babywalk} exist for visually grounded language guided navigation. Thomason et al.~\cite{jesse1,cvdn} worked on human-robot for dialog-driven tasks. However, these approaches have neglected the affective states of the language commands. 

  


\subsection{Reinforcement Learning in Social Robotics}
Reinforcement learning~\cite{suttonrl} has been used in social robotics to learn desired behavior through interaction with humans. Akalin et al.~\cite{RLsoc1} present a review of social robotics work using RL. Weber et al.~\cite{humorrl} used RL to adapt to a human companion's sense of humor from vocal and visual cues of laughter and smiling. Papaioannou et al.~\cite{chatrl} used RL to increase customer engagement on a robot guide. Along with learning from demonstration with social reward signals, recent works implement intrinsic rewards for a robot tasked to perform handshakes to learn more human-like behavior~\cite{intrinsicrl, multimodalrl}. Homeostasis based reward formulations have also been explored by Alvaro et al.~\cite{homeo1,homeo2,homeo3}. In our method, we formulate a social reward to encourage interactions and an intrinsic exploration strategy to reach the goal.

 


\section{Overview}
\label{sec: Overview}
\begin{figure}[t!]
    \centering
    \includegraphics[width=1\linewidth]{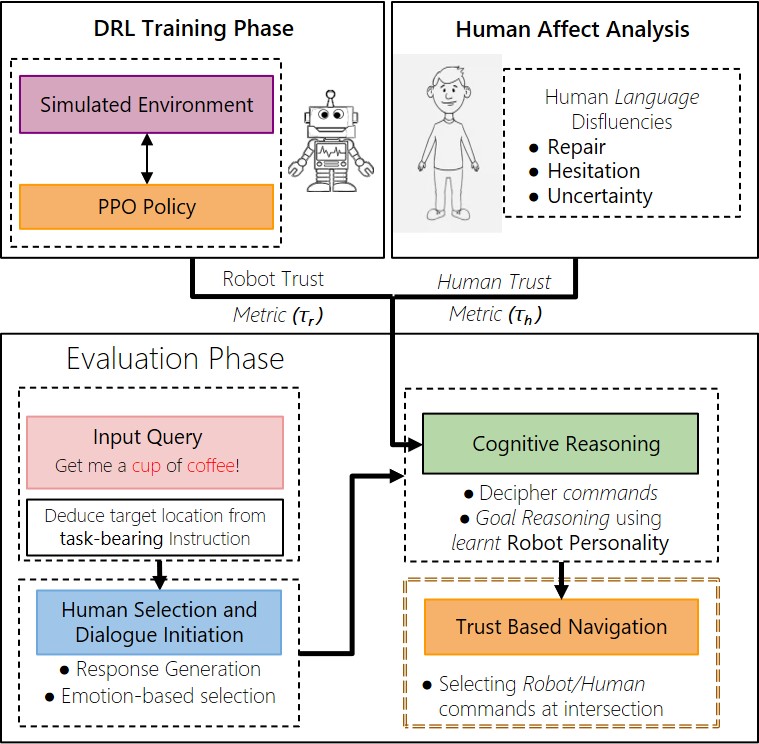}
    \caption{\textit{\textbf{Overview of our approach:} We train a DRL policy to learn how much trust the robot must place on humans in an arbitrary environment. This \textit{robot trust metric} ($\tau_{r}$) drives our trust based navigation approach. Our policy also learns to produce reactive navigational actions. During evaluation, given an input task-bearing query, we first deduce a target location (\textit{kitchen} in this case). In order to obtain directions towards the goal, the robot chooses a socially-aware human in the environment and initiates a dialogue for directions. Then, a cognitive reasoning step evaluates the \textit{human trust metric} ($\tau_{h}$) and $\tau_{r}$ learned by our policy, to decide what action the robot must take at the next gateway.
    A \textbf{\textit{gullible}} robot has a low $\tau_{r}$ and would would choose to trust all navigation instructions irrespective of $\tau_{h}$. A \textbf{\textit{skeptical}} robot on the other hand would reject all human guidance, and navigate only according to the policy. Our DRL model is tasked with producing an appropriate $\tau_{r}$ that can gauge the average $\tau_{h}$ of the environment.} \vspace{-0.5cm}}
    \label{overview}
\end{figure}

Figure \ref{overview} outlines our approach in two phases. We first train a DRL policy to optimize navigation towards a goal and learn a \textit{robot trust metric ($\tau_{r}$)}. This quantifies a level of trust to place on humans in a particular scene. Using \textit{affective} cues from human speech, we obtain a \textit{human trust metric ($\tau_{h}$)}. In the second phase, we utilize the $\tau_{r}$ and $\tau_{h}$ to reason about directional commands and navigate in an optimal, time-efficient manner.
\subsection{Notations}
We use the following notations throughout the paper.
\begin{enumerate}[nolistsep]
    \item $S$ represents symbols obtained from natural language instructions, after \textit{symbolization}.
    \item $A$ refers to affective features extracted from a human. In psychology, \textit{affects} refer to the expressed emotions or feelings, both in a verbal and non-verbal sense. In our work, we use verbal affective cues.
    \item $\tau_{h}$ is the trust metric of a human. It tells us how confident humans are in the guidance they give. We compute this as a function of the affective features obtained from the spoken language, given by $\tau_{h} = \Pi_{i=0}^{n}(A_{i})$, where $A_{i}$ represents a probabilistic score of detecting the $i^{\text{th}}$ affect.
    \item $\tau_{r}$ is the trust metric of the robot and is used to interpret if the robot can trust humans in the environment or not based on $\tau_{h}$. This also defines the robot's personality as \textbf{gullible} or \textbf{skeptical}. We learn this using our DRL policy  $\pi_{\theta}(a|o)$, where $\theta$ represents the optimal parameters for the policy $\pi$ which maps actions $a$ to observations $o$. $\tau_{r}$ is part of the predicted action space $a$.
\end{enumerate}

\subsection{Training Phase}
We train our RL policy in a simulated corridor like environment with virtual humans. During each training episode, the robot is tasked with reaching the target using directional information obtained from human interactions. States are updated by an \textit{event-triggered} timestep, which occurs whenever the robot detects either a \textit{gateway} or a human. The observation space consists of language commands at every state, their respective confidence scores, and a count of interactions had and gateways visited so far. The policy then chooses from a set of actions corresponding to orientation changes and interactions.

The policy also outputs a \textit{robot trust metric} $\tau_{r}$ at each state to gauge human trust $\tau_{h}$ in the environment. These metrics are used to define a ``social reward" while training our policy, which ultimately drives the type of personality that the robot must assume in a given environment.
The trained DRL policy not only learns to follow human guidance to navigate towards the target but also learns what guidance to trust in order to navigate in the most time-efficient manner. 
\subsection{Evaluation Phase}



Given an input task-bearing instruction \cite{jspeech}, the target location is deduced, and we begin our trust-driven navigation pipeline. The trained RL policy decides the robot's actions when it detects humans or encounters gateways in the environment.
When the robot detects and chooses to interact with humans for navigational guidance, it anticipates \textit{executable} instructions that it can convert to directional \textit{language symbols}.
For example, a command like ``\textit{Go straight and take the second left}" is reducible to ``\textit{UUL}" in terms of \textit{language symbols}. These symbols correspond to the direction that the robot must take (straight, left, or right) when it reaches a \textit{gateway} \cite{interdetect1, interdetect2} in the environment.



Along with the extracted language symbols, we also obtain verbal \textit{affective} cues ($\mathbf{A}$) from both speech and the transcribed text to identify parts of the instruction that sound confident. These cues (disfluency in speech, uncertainty in transcription) give us context about the speaker's credibility in his or her ability to guide the robot \cite{credible1}.
$\mathbf{A}$ helps determine a probabilistic trust metric $\tau_{h}$ for each of the individual language symbols obtained. This tells us how confident the human is on each of the directions they give, which can be thought of as quantifying the human's sense of direction.
For example, consider the guidance \textit{``Go straight at the intersection and take the next left, and umm no-no take a right}. Here, the human clearly sounds confident about going \textit{straight (U)} at the intersection, but seems unsure about whether to take a right (R) or a left (L) afterwards. Our algorithm identifies the unreliable nature of this guidance and automatically assigns a relatively high confidence score to the \textit{straight} or \textit{(U)} language symbol but gives a much lower confidence score to the other directions. This score is part of the observation space of our policy.

Then, at each gateway, the policy determines the action that the robot needs to take in order for it to reach the goal in an optimized manner. \textit{Optimality} for our task here is defined as minimizing the amount of time and interactions needed to reach the goal.

Our learned policy $\pi$ predicts $\tau_{r}$, which is used to define its personality in terms of \textbf{skeptical} or \textbf{gullible} nature.
An overtly \textbf{\textit{skeptical}} robot has a low $\tau_{r}$, which drives it to ask guidance from a new bystander whenever the opportunity arises. This causes navigational delays due to excessive interaction time.
On the other hand, an overtly \textbf{\textit{gullible}} robot places high trust on all human guidance, which causes it to fully execute every single directional command it receives, not trusting its own intelligence. This is not advantageous either, as repeated unreliable guidance from humans can lead to lengthy robot navigation routes, which sometimes cause it not to reach the goal at all.

Our policy learns to gauge the $\tau_{h}$ of the environment to produce a robot navigation scheme that is intelligent enough to seek guidance from a human when in doubt, but that is also not too naive in trusting all the information it gets. 
To this end, we carry out experiments to train our DRL model in a simulated environment with varying parameters (variance in $\tau_{h}$, distance to goal) and showcase results.

\section{Method}
\label{sec: Method}

\subsection{Affective Analysis}

\subsubsection{\textit{Lang2Sym} Dataset Preparation}
\label{langgen}
Symbolization refers to the conversion of natural language instructions to suitable directional symbols ($S$) for navigation. In order to perform this task, we first collect a \textit{Lang2Sym} dataset containing natural language instructions for navigational guidance.
Annotators of this dataset were given a floor plan with a robot positioned at a random starting location and were asked to give language-based guidance to a given target location. We then parse these instructions through a rule-based symbolizer~\cite{matuszek2010following}, which uses the order and sequence of \textit{directional} words such as forward, straight, left, right, etc. to extract symbols from them. Table \ref{tab:rules} provides an overview of the different types of cases that we encounter and the generated language symbols.



\begin{table}[h!]
\centering
\caption {Rules for Symbolization} \label{tab:rules}
    \begin{tabular}{|p{5.5cm}| p{2.5cm}|}
    \hline   \textbf{Example Cases} & \textbf{Symbols Generated} \\  \hline
    ``\textit{Go \underline{straight and take the second left}}" & U, U, L \\  \hline
   ``\textit{Go forward till the \underline{end of the corridor} and turn left}" & 
   U, U,.., L\\  \hline
    ``\textit{\underline{Turn left} and go straight}" & L U\\  \hline
    ``\textit{Go straight and take a right towards the kitchen \underline{on your right}}" & U,  R\\  \hline

    \end{tabular}
  \vspace*{-0.05in}   
\end{table}

Ultimately, we obtain four types of symbols - $U$ for Upward, $D$ for Downward. $L$ for Left, and $R$ for Right. This \textit{Lang2Sym} dataset has been released and can be found on our website.

\subsubsection{The Human Trust Metric ($\mathbf{\tau_{h}}$)}
\label{confidence_score}
Humans are capable of using a variety of verbal \textit{affective} cues to construct their own ideas of credibility and trustworthiness~\cite{trustaff1} of others while interacting with them. This, in turn, affects their decision-making skills~\cite{trustaff2}.

\textit{Disfluency} in literature has been described by several names, including hesitation, disturbance, fragmentation, hemming, and hawing~\cite{disfluency-2}. In our work, we define disfluency as described in~\cite{disfluency-2}, where it refers to all the different types of alterations in speech.

When humans need to give directional guidance, often is the case when people are unable to describe goal locations accurately. We look into three cases of disfluency in particular and assign probabilistic values for how confident the human sounded in their speech. These are defined as follows:
\begin{itemize}
    \item \textbf{Speech Repair ($\mathbf{A_{rep}}$)}-
    This happens when a human \textit{corrects} or \textit{repairs} their own speech after fumbling.
    
    For example, a human might say, ``Take a right \underline{uhh} I mean take a left and go straight". Parsing this speech without taking disfluency into consideration would give us the language symbols $R, L, U$, while what the human wanted to convey was $L, U$. We detect these using an approach described by Jamshid et al.~\cite{speechdis}, which gives us the part of speech on which the human fumbled. 
    \item \textbf{Hesitation Signals ($\mathbf{A_{hes}}$)}- This type of speech disfluency occurs when a human hesitates or pauses while speaking \cite{hes}. For example, a human might say, ``Go straight and \underline{\textit{err}} take the second left". While the human does sound confident about going straight here, the hesitation right afterward might indicate that they were not confident about which left to take.
    \item \textbf{Language Uncertainty ($\mathbf{A_{unc}}$)}- Certain keywords such as ``\textit{probably, may, might, I think}" in the transcribed text indicate a lack of confidence in the path described. While the human might sound confident in his speech, these uncertainties in the transcribed text need to be accounted for by our symbol generator.
\end{itemize}

These aberrations in speech and the transcribed text give us a measure of how ``trustworthy" a human's directional guidance might be. In order to account for these, we assign a probabilistic score to each of the symbols generated whenever any of these aberrations get detected. In our experiments, these values are determined from the results presented in \cite{confidencescore2,confidencescore3,confidencescore4} and are set as $A_{unc} = 0.5$, $A_{rep} = 0.75$, and $A_{hes} = 0.5$, and apply them in accordance to where the aberrations were detected.
The net Affective feature $A_{net}$ is then given by $\mathbf{A_{net}} = \mathbf{A_{rep}} \times \mathbf{A_{hes}}\times \mathbf{A_{unc}}$.



The human trust metric $\tau_{h}$ is the average of probability across all symbols, given by:
\begin{align}
    \tau_{h} = \frac{A_{net}}{\sum_{i=0}^{i}{i}}, \quad i\neq0
\end{align}
where $i$ is the index value representing each symbol.
\subsection{Simulation Setup}
In our simulations, we model multiple humans $\mathcal{H}$ walking in the environment along with a robot. Each human is mathematically modelled using the following equation:
\begin{align}
    \mathcal{H} = \{\mathbb{M}, \:\: S[i], \:\: \tau_{h}[i]\}, \:\: i \in [0, l_{max}]
    \label{eq:human}
\end{align}
where $S$ are the extracted language symbols and $\tau_{h}$ is the corresponding confidence score as defined in \ref{langgen} and \ref{confidence_score} respectively. $\mathbb{M}$ represents the \textit{mental map} that a human has of the environment. $l_{max}$ is the maximum length of $S$.

\subsubsection{Human Modeling}
\label{generation}
We generate language symbols $S$ for simulation by performing permutations on a graph-based search from the current position of the human to the goal location in their mental map $\mathbb{M}$. This map is an undirected graph with nodes representing the gateway points in the environment.
The trust metric $\tau_{h}$ for each human in a simulated environment is calculated using a 2D Gaussian distribution centered around the goal region given by
\begin{align}
      G(x, y) &= {\frac{1}{2\pi\sigma^{2}}\exp({-\frac{(x-x_{g})^{2}+(y-y_{g})^{2}}{2\sigma^{2}}})} 
      \label{eq:gauss}
\end{align}
where ($x_{g}$ , $y_{g}$) and ($x$, $y$) are the coordinates of the goal and start position respectively, and $\sigma$ is the standard deviation. This process is illustrated in figure \ref{fig:probability} below.

\begin{figure}[h]
\centering
\includegraphics[width=\linewidth]{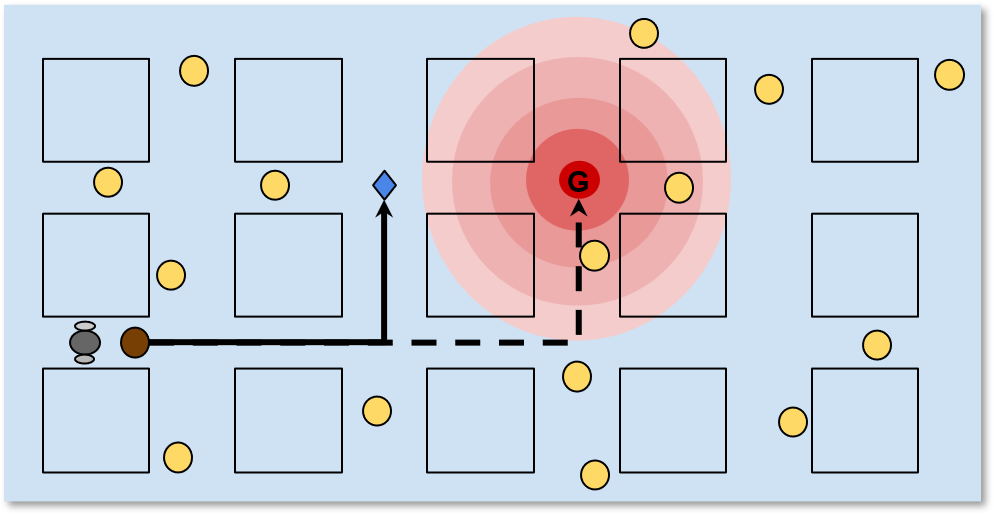}
\caption{\textit{\textbf{Human Modeling}: People (yellow) randomly move about in the environment. When the robot (gray) decides to interact with someone (brown), it receives language symbols $S$ and an associated confidence score $\mathbf{\tau_{h}}$ for each symbol. Both these values are computed using the mental map $\mathbb{M}$ of the human, which is a graph whose nodes are gateway points. The goal, according to \textit{brown}, is shown by the blue diamond ($[R, U]$). Notice this is different from the actual goal \textbf{G} ($[R, R, U]$). We model this artifact intentionally by considering neighboring nodes to \textbf{G} in $\mathbb{M}$, which adds to the randomness of the generated humans. To model $\mathbf{\tau_{h}}$, we use a 2D Gaussian described in equation \ref{eq:gauss}.} \vspace{-0.5cm}}
\label{fig:probability}
\end{figure}

\subsection{Training our Reinforcement Learning Policy}
\label{DRL}
We draw inspiration from human behavior in asking for guidance when placed in a new environment to motivate our DRL approach. Over time, through repeated guidance, humans learn more and more about the environment and can navigate easily. With the help of the feedback given, they are able to \textit{reinforce} upon whose guidance can be trusted based on the environment. Our DRL model tries to emulate this behavior in the same way on a robot, choosing what guidance to trust and what not to, by repeatedly interacting with the environment and updating its policy.

\subsubsection{DRL Model Architecture}
\begin{figure}[t]
  \centering
  \includegraphics[width = \linewidth]{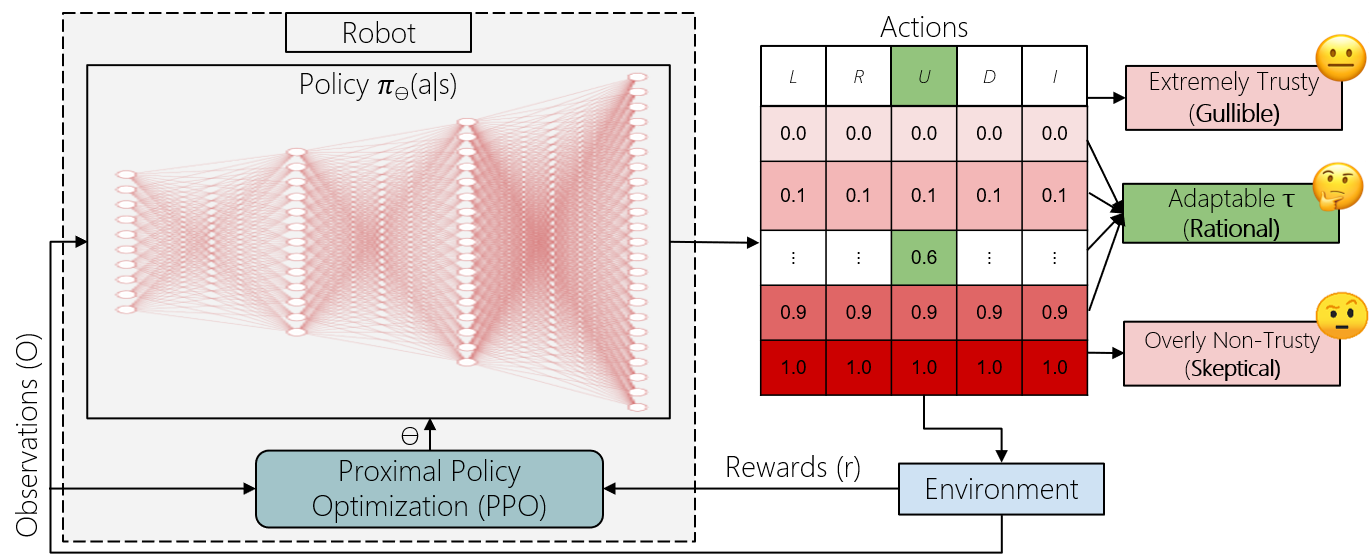}
  \caption{\textit{\textbf{Overview of DRL pipeline:} We use a PPO model to train our policy. A new state gets triggered via an event-based timestep, which occurs whenever the robot detects either a gateway or a human. The observation space consists of numerical representations of language guidance $S$ received along with the respective human confidence scores $\tau_{h}$. The policy has an MLP architecture that predicts a value over a 2D action space (flattened to 1D) consisting of \textit{reactive} commands (Directions $L, R, U, D$ + Interaction $I$) and the robot trust metric $\tau_{r}$. The policy must learn to predict optimal values of $\tau_{r}$ in order to navigate the environment for the maximum reward. Here, the reactive command selected is $U$, asking the robot to go up at the next intersection, and $\tau_{r}$ predicted is 0.6. \vspace{-0.5cm}}}
  \label{MLP}
\end{figure}

We use a policy gradient approach, Proximal Policy Optimization (PPO) \cite{ppo}, for training our model. This is an \textit{on-policy} approach, which means that the actions are taken based on the current approximation of the optimal policy and not greedily on a sub-optimal policy. In our case, as our agent needs to explore the environment judiciously, this approach allows for faster convergence. Figure \ref{MLP} illustrates our DRL training pipeline. Our policy learns to navigate to the goal optimally using human guidance by predicting a $\tau_{r}$ at every state.

\subsubsection{State and Action Spaces}
Our simulated environment consists of several rooms and humans, and thus several \textit{decision-making} opportunities where the robot has to decide which direction to take.

Humans $\mathcal{H}^{n} \:\: | \:\: n \in [h_{min}, h_{max}]$ are generated randomly at each episode, with language symbol sets $\{S[i]\:|\:i\in[0,\:\: l_{max}]\}$ and corresponding \textit{confidence scores} $\{\tau_{h}[i]\:|\:i\in[0,\:\: l_{max}]\}$ using the algorithms described in \ref{generation}. 


To speed up simulations, we model humans as low poly cuboids that move in the environment. They are modeled to have a constant velocity chosen randomly between  $0.8 m/s$ to $1.2 m/s$~\cite{walking}, and move vertically or horizontally from one end of a corridor to the other.


Our observation space is defined as follows:
\begin{align}
    \mathcal{O}=\{S, \tau_{h}, N_{I}, N_{G}, D_{h}, D_{t}\}
\end{align}
Here, $S$ and $\tau_{h}$ are arrays of length $l_{max}$ containing the numerical representations of language symbols and the corresponding confidence scores; $D_{h}$ and $D_{t}$ are the boolean parameters corresponding to detecting the human and detecting the target respectively; $N_{I}$ and $N_{G}$ are a count of the number of interactions and the number of gateways detected. Until an instruction is received, the $S$ and $\tau_{h}$ remain empty.

The timestep for our RL algorithm is event based, and a new state is triggered whenever a gateway or a human is detected. The default action of the robot between states is forward movement with respect to its orientation. Then once a timestep gets triggered, the policy outputs an action from the following two dimensional action space:
\begin{align}
    \mathcal{A} = \overbrace{\{L, R, U, D, I\}}^\text{Reactive Commands} \times \underbrace{\{0, \dots, 1\}}_\text{Discrete $\tau_{r}$ values}
\end{align}

\begin{figure*}[t!]
    \centering
      \includegraphics[width = \linewidth]{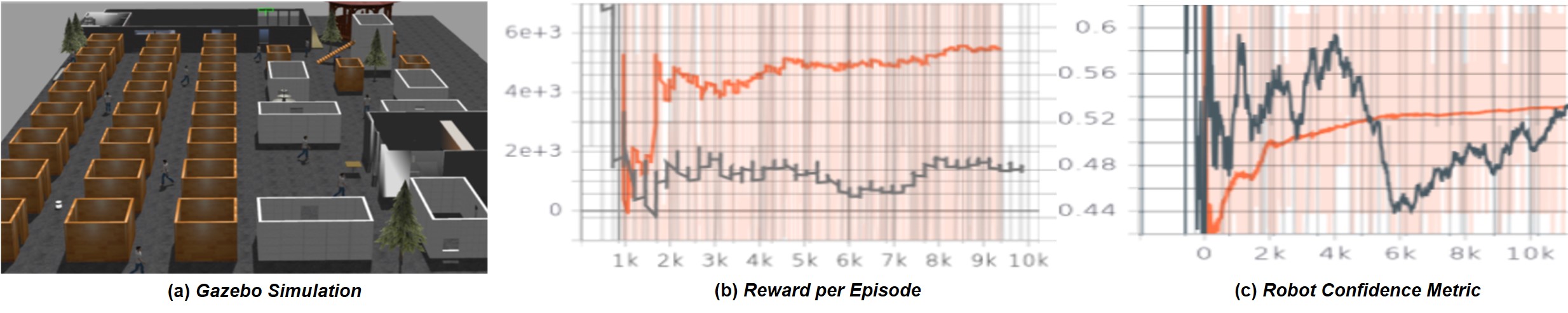}
    \caption{ \textit{ (a) We train our model on a variety of simulation scenarios that include intersections and walking humans. (b) A comparison of the episode rewards of a random exploration strategy (black) vs our trained policy. Observe the upward trend in obtaining rewards when the robot learns to interact with people and take guidance from them. (c) A comparison of the $\tau_{r}$ values of the exploration graph vs. the outcome of our policy. The exploration graph behaves in a haphazard manner, while our policy converges to maximize its reward by setting $\tau_{r}$ to the $\tau_{h}$ of the environment. }}
    \label{fig:experiments}
    \vspace{-0.6cm}
\end{figure*}

Here, $L, R, U,$, and $D$ are directional actions that orient the robot in a particular direction, while $I$ is an interactive action, where the robot chooses to interact and update $S$ and $\tau_{h}$. Note that the robot needs to learn to interact only when it detects a human in its state space ($D_{h} = True$). Thus, interactive actions at gateways will not have any effect.

At each timestep, the policy also predicts a robot \textit{trust metric} $\tau_{r}$ along with the action, which determines what level of trust the robot has on humans' instructions in the environment. 
The value of $\tau_{r}$ is discretized between $0$ and $1$ by a factor of $10$, and thus our action space has a dimensionality of $\mathbb{R}^{5\times10}$, which is flattened to $\mathbb{R}^{50}$ to match the dimension of $\mathcal{O}$.

\subsubsection{Reward Formulation}
We formulate four types of rewards to motivate the robot to find the most optimal path towards the goal.

\begin{enumerate}[wide, labelwidth=!, labelindent=0pt]
    \itemsep 0.5em
    \item \textbf{Interaction Penalty} ($\mathbf{r_{i}}$):
    This is a negative reward to penalize the robot when it is expected to interact with humans, but chooses otherwise. It is defined as:
        \begin{align}
        \mathbf{r_{i}} = 
        \begin{dcases}
    r_{nointer}, &if \:\: D_{h} = True \:\: \& \:\: N_{i}<I_{min} \:\: \& \:\: \mathcal{A} \neq I\\
    r_{wronginter}, & if \:\: D_{h} = False \:\: \&\:\: \mathcal{A} = I \\
    0, & otherwise
        \end{dcases} \notag
        \end{align}
    
    $r_{nointer}$ occurs when the robot has the opportunity to interact (i.e., detect human $D_{h}=True$), has so far interacted too little (i.e., $N_{I}<I_{min}$), and is still choosing not to interact ($\mathcal{A} \neq I$).
    
    On the other hand, if the robot does not detect a human ($D_{h}=False$) and chooses to interact ($\mathcal{A} = I$), it needs to be penalized.
    
    \item \textbf{Target Reward} ($\mathbf{r_{t}}$):
    We give a positive reward when the robot reaches the desired target location, and a negative reward otherwise. It is formulated as: 
        \begin{align}
        \mathbf{r_{t}} = 
        \begin{dcases}
        r_{reached}, &if \:\: D_{T} = True \\
        0, & otherwise
        \end{dcases} \notag
        \end{align}
    \item \textbf{Delay Penalty} ($\mathbf{r_{g}}$):
    We set a reward on the time that the robot takes to reach a goal, based on the number of gateways it passes ($N_{G}$). This is formulated as: 
        \begin{align}
        \mathbf{r_{g}} = 
        \begin{dcases}
        r_{gmin}, &if \:\: N_{G} = N_{Gmin} \\
        -w_{n}(N_{G}-N_{Gmin}), & if \:\: N_{G} > N_{Gmin}\\
        0, & otherwise
        \end{dcases} \notag
        \end{align}
    Here, the weight multiplier $w_{n}$ translates to a negative penalty for each extra gateway that the robot passes through, beyond the minimum number $N_{Gmin}$. This is the shortest path to goal, mentioned in section \ref{generation}.
    \item \textbf{Social Reward} ($\mathbf{r_{s}}$):
    In order to encourage the robot to follow human commands at a particular timestep, we delegate a social reward each time it interacts with someone. 
        \begin{align}
        \mathbf{r_{s}} &= 
        \begin{dcases}
        r_{follow} + r_{trust}, &if \:\: \mathcal{A}=S[t]\\
        0, & otherwise
        \end{dcases} \notag \\
        r_{trust} &=  \begin{dcases} 
        -w_{o} \left|(\tau_{h}[t] - \tau_{r} )\right| , &if \:\: \tau_{h}][t] \neq \tau_{r}\\
        r_{optimal}, & otherwise
        \end{dcases} \notag
        \end{align}
    Here, $r_{follow}$ is a constant reward for choosing to follow human commands. $r_{trust}$ is a weighted reward on the difference between the trust metrics. The optimal condition for $r_{optimal}$ occurs when $\tau_{r}$ equals $\tau_{h}$.
\end{enumerate}

The total reward $\mathbf{r_{total}}$ here is a summation of all the rewards combined given by $\mathbf{r_{total}} = \mathbf{r_{i}} + \mathbf{r_{t}} + \mathbf{r_{g}} + \mathbf{r_{s}}$. For experimental details, please refer to our website.

Our rewards values are modeled based on observations made in three stages of our experimentation. First, we gave a high reward value to $r_{nointer}$, so as to make the robot interact with humans in order to gain a greater reward. This caused the model to only select the interactive action ($\mathcal{A}=I$) at each step. In order to rectify this, we introduced $r_{wronginter}$.

In the second stage, the robot learns when to interact with humans to increase its reward. However, it would not follow human guidance, after interacting with them. To tackle this, we introduced $r_{follow}$ as a constant social reward for following human guidance. Setting $r_{follow}$ to a high value caused the robot to behave in a gullible manner, not accounting for $\tau_{h}$.

In our final stage of reward modeling, we addressed this gullibility via a trust reward $r_{trust}$. When the robot blindly follows human commands without choosing the optimal $\tau_{r}$ to match $\tau_{h}$, it is penalized. Otherwise, it is rewarded with a positive $r_{optimal}$.



\section{Experiments and Results}
\subsection{Simulation Results}

We train our model in simulated environments consisting of randomly spawned humans $\mathcal{H}$, having a spread of confidence scores obtained using the modeling design described in \ref{generation}. $l_{max}$ for the symbols $S$ is set to $5$ in our experiments. At each episode, the robot and goal are spawned at a fixed location in the environment (See fig. \ref{fig:experiments}). New states are triggered whenever the robot detects a human or a gateway, identified using a simulated 2D Lidar and RGBD camera. We train our policy on an Nvidia GeForce RTX 2080Ti GPU with a learning rate of $0.1$, an episode length of $15$ steps, for 10K steps.

Figure \ref{fig:experiments} (b) and (c) show comparisons between an exploratory robot that chooses random actions (black) at each state versus our trained policy (orange). The upward trend in the rewards graph (b) generated by our policy converges at a maximum reward value of around 5K in 10K steps. In contrast, the random exploratory actions outputs a much lower sub-optimal reward curve. A similar trend of convergence can be seen in the robot trust metric output $\tau_{r}$. While our learnt $\tau_{r}$ converges at around 0.5, the random policy does not converge. The convergence value of $\tau_{r}$ here represents the average $\tau_{h}$ of the environment.


\subsection{Real World Experiments}
As our observation space contains low dimensional data of language symbols $S$, transferring the learned model onto a real-world environment required minimal changes. We used an object detector and a distance-based gateway detection scheme for identifying people and intersections, respectively. $\tau_{h}$ was estimated using the \textit{affects} described in section \ref{confidence_score}.

We devised two experiments in our indoor lab environment to evaluate the learned $\tau_{r}$. In the first experiment, we ask humans to give trusty directions (high $\tau_{h}$), and in the other, we ask them to give non-trusty directions (low $\tau_{h}$). We then associate the learned skeptical and gullible models to each of them and perform a comparison on the performance of both. Please refer to the supplementary video and website for more details.

\section{Conclusion}
\label{sec:Conclusion}

We present a novel \textit{trust-driven} approach for optimizing language guided navigation in an unseen environment. Our method makes use of \textit{affective} features computed from human language combined with a Deep Reinforcement Learning policy that learns to make intelligent decisions about the guidance given. Apart from learning to follow human guidance to reach the target location, our policy also learns a robot trust metric $\tau_{r}$ which gauges the overall confidence level $\tau_{h}$ of humans in the environment. We also release our \textit{Lang2Sym} dataset, used to compute language symbols from human language instructions for navigation.

Our model's limitation is that it relies on noisy audio cues to compute disfluency and can lead to incorrect computation of trust. In the future we would like to compute a multi-modal trust affect using a combination of verbal and non-verbal cues (like gaits and gestures). We also plan to integrate the valence-arousal-dominance emotion model to better model human and robot trust metrics.

                                  
{\small
\bibliographystyle{IEEEtran}
\bibliography{refs}

}

\end{document}